\documentclass[journal]{IEEEtran}
\usepackage{graphicx} 
\usepackage{pgfplots}
\usepackage{booktabs}
\usepackage{adjustbox}
\usepackage{subcaption}
\usepackage{cite}
\usepackage{hyperref}

\usepackage{amsmath,amsfonts,amssymb}
\title{Dual-Stream Attention Transformers for Sewer Defect Classification}
\author{ Abdullah Al Redwan Newaz$^{1,2, *}$, Mahdi Abdeldguerfi$^{1,2}$, Kendall N. Niles$^{3}$, and Joe Tom$^{3}$
\vspace{-15pt}

\thanks{$^{1}$ Canizaro Livingston Gulf States Center for Environmental Informatics, The University of New Orleans, New Orleans, LA 70148, USA.}
\thanks{$^{2}$ Department of Computer Science, The University of New Orleans, New Orleans, LA 70148, USA. }
\thanks{$^{3}$ US Army Corps of Engineers, Vicksburg District, MS 39183 USA.}

\thanks{*Corresponding author: Abdullah Al Redwan Newaz.
Address: 2000 Lakeshore Dr,
Department of Computer Science,
University of New Orleans, New Orleans, LA 70148, USA. Email:
\quad{\tt\small aredwann@uno.edu}}
}

\begin{document}

\maketitle

\begin{abstract}
We propose a dual-stream multi-scale vision transformer (DS-MSHViT) architecture that processes RGB and optical flow inputs for efficient sewer defect classification.
Unlike existing methods that combine the predictions of two separate networks trained on each modality, we jointly train a single network with two branches for RGB and motion.
Our key idea is to use self-attention regularization to harness the complementary strengths of the RGB and motion streams. The motion stream alone struggles to generate accurate attention maps, as motion images lack the rich visual features present in RGB images.
To facilitate this, we introduce an attention consistency loss between the dual streams. By leveraging motion cues through a self-attention regularizer, we align and enhance RGB attention maps, enabling the network to concentrate on pertinent input regions.
We evaluate our data on a public dataset as well as cross-validate our model performance in a novel dataset.
Our method outperforms existing models that utilize either convolutional neural networks (CNNs) or multi-scale hybrid vision transformers (MSHViTs) without employing attention regularization between the two streams.
\end{abstract}


\section{Introduction}

\IEEEPARstart{A}{utomated} sewer inspection robots frequently require a thorough grasp of their surroundings through the use of numerous sensory inputs. These inputs can include visual data (RGB images)~\cite{8013727}, motion data (changes in the environment over time)~\cite{8038234}, and potentially other types of sensory data. By learning from these multiple modalities, a more comprehensive understanding of defects in the sewerage environment can be obtained.

Convolutional neural networks (CNNs) have typically been the predominant method for automated sewer inspection~\cite{panta2023iterlunet, panta2022pixel, kuchi2021machine, kuchi2020levee, kuchi2019machine}. While a CNN efficiently models local spatial semantics within the image, it struggles to model non-local spatial semantics. Recent work Multi-Scale Hybrid Vision Transformer (MSHViT) explicitly models non-local spatial semantics across scales leading to a relative
improvement compared to using just the CNN backbone for sewer image classification~\cite{haurum2022multi}. However, none of these methods utilize RGB images and optical flow for efficient sewer defect classification. 

Ensemble methods~\cite{islam2022pedestrian, islam2021pedestrian} combine outputs from networks trained separately on RGB and motion data. However, neither these approaches can directly improve the individual network performances nor do they account for the inherent differences in information content between modalities.
For example, motion images from various sensors like LiDAR and cameras lack the rich visual features present in RGB images. RGB contains color, texture, shape, and other informative details that aid recognition but affected by illumination conditions and motion blurs. In contrast, motion only encodes movement patterns without a broader visual context. As a result, models trained solely on motion often have deficient features compared to RGB networks, causing more false positive predictions.
Additionally, concatenating multiple networks increases overall architecture complexity, sacrificing runtime efficiency.

Detection and tracking tasks often learn from sequential images using correlation layers to assess local feature similarity between consecutive frames across CNN feature maps~\cite{yin2021center, feichtenhofer2017detect}.
Though self-attention transformers now surpass CNNs in focusing on relevant regions, leveraging self-attention across different modalities is still unexplored.

We address this by introducing an attention regularization loss that aligns and enhances the attention maps through shared learning between the RGB and motion streams. This allows better utilization of available information from both sources.
Unlike ensemble methods, our joint training approach penalizes the motion network for attending to less informative RGB regions. By regularizing motion attention with guidance from richer visual features, we overcome the limitations of training the streams independently. The RGB cues provide context to refine motion attention, improving accuracy without separate complex architecture.

We train dual-stream multi-scale vision transformers (DS-MSHViT) on both RGB and motion data. The inputs are fed through parallel transformer streams. An attention consistency loss then optimizes the alignment between the attention maps from both streams. After training on the multimodal dataset, the RGB network performance is significantly improved. Only the RGB branch is then used for prediction on the validation and test sets.
This approach allows us to leverage the richer features from RGB to guide and stabilize the attention for motion, overcoming the limitation of sparse motion inputs. As a result, our dual-stream attention transformer achieves significantly improved accuracy compared to standard ensemble methods by leveraging inter-stream attention regularization.
The contributions of this work are:
\begin{enumerate}
    \item The development of a dual-stream multi-scale vision transformer (DS-MSHViT) architecture to effectively utilize multimodal RGB and motion data for accurate prediction of sewer deficiencies.
    \item The proposed DS-MSHViT can leverage sequential image inputs to accurately detect deficiencies across consecutive frames.
    \item The development of a novel validation dataset comprising 50,000 multi-label annotations on sequential video frames. This enables rigorous benchmarking of model performance for defect detection across consecutive images.
    \item Extensive evaluations that demonstrate a clear advantage of the proposed DS-MSHViT over baseline MSHViT models, with improved performance across multiple metrics. 
\end{enumerate}

The remainder of the paper is organized as follows:
Section~\ref{Section:related} reviews existing works on sewer defect classification techniques; Section~\ref{Section:Proposed} introduces the proposed DS-MSHViT architecture for an efficient combination of RGB and motion inputs; Section~\ref{Section:Experiment} presents experimental results benchmarking baseline models and their classification performance; and finally, Section~\ref{Section:Conclusion} provides concluding remarks and a summary of key contributions.
\section{Related Work}\label{Section:related}
This section provides a brief literature review covering vision-based defect inspection, recent advances in lightweight vision transformers, and optical flow algorithms.
\subsection{Vision-based Defect Inspection}
Sewer pipe inspection has long utilized automated systems with tailored image processing for detecting defects \cite{duran2002automated, li2022vision}. However, deep learning breakthroughs have recently led to major advancements in analysis techniques for automated sewer inspection \cite{kumar2020deep, ji2021determination, tan2021automatic, dang2022defecttr, biswas2023sewer}. Now, deep learning models are the primary focus in this domain instead of traditional computer vision pipelines. Specifically, CNN classifiers are often used for defect classification, as they can automatically extract features from annotated data \cite{haurum2020water, ji2020measurement, kuchi2021machine, kuchi2020levee, kuchi2019machine, tao2022cafen}. However, training these neural networks requires large annotated datasets, which are scarce. The extreme class imbalance in available data makes training very challenging, often leading to overfitting. Some work explores semi-supervised and unsupervised learning to develop defect detection systems without full annotation. For example, autoencoders can model water levels in sewers with sparse labels \cite{plana2022autoencoders}. Cycle-GAN and YOLO v5 have been used for petrochemical pipeline deficiencies \cite{chen2022automatic}. Recent emphasis is also on extending CNNs to pixel-level semantic segmentation \cite{wang2020unified, zhou2022automatic,li2022robust, panta2022pixel, panta2023iterlunet}.

Multilabel classification presents an inherent challenge for extracting discriminative features for different defect classes from the relevant image regions~\cite{hu2023toward, zhao2023towards}. While CNN-based classifiers benefit from data-driven feature extraction, the ambiguity of sewer defects in the feature space makes multilabel defect classification difficult. Recent work uses a multi-scale hybrid vision transformer to address this ambiguity~\cite{haurum2022multi}. Inspired by this, we develop a dual-stream multi-scale hybrid vision transformer for robust sewer inspection using multi-modal transformer learning.


\subsection{Lightweight Vision Transformers}
Vision transformers (ViTs) have gained significant research interest lately owing to their ability to learn global representations using self-attention, making them well-suited for computer vision tasks~\cite{9674785}. For sewer inspection applications, lightweight ViT models are preferable since they could be deployed on battery-powered robots with limited computational capability. Lightvit \cite{huang2022lightvit} balances accuracy and efficiency using pure transformer blocks without convolution.

A popular technique to reduce computation cost for transformers is developing hybrid models combining CNN and ViT architectures \cite{zhao2022lightweight, li2022using, mehta2021mobilevit}. Another idea uses efficient tokenizers like Sinkhorn or joint token pruning and squeezing (TPS) to compress ViTs with higher efficiency. However, none of these methods utilize attention regularization during training to improve performance while using a single lightweight ViT at test time. The proposed DS-MSHViT is the first method that utilizes this idea of attention regularization during training for improved efficiency and performance with a single lightweight ViT architecture at test time.

\subsection{Optical Flow Estimator}
Videos of sewerage systems are commonly recorded using mobile robots equipped with closed-circuit television (CCTV) cameras. When assessing the condition of sewers using such video footage, optical flow estimation emerges as a natural choice. However, CNN-based optical flow estimators are prone to various noise interferences during the computation of pixel-wise 2D motions between successive video frames. These estimators rely heavily on local features, leading to challenges in matching corresponding pixels, as evidenced in previous research~\cite{hui2018liteflownet}.

In the realm of computer vision, transformer-based approaches have demonstrated superior performance in optical flow compared to their CNN-based counterparts, as shown in recent studies~\cite{luo2022learning, zhao2022global, sui2022craft, zheng2022dip, jeong2022imposing, sun2022skflow, zhao2022global}. While transformer-based methods have achieved state-of-the-art results on various computer vision tasks, none of the existing methodologies have been specifically tailored and applied to the challenging problem of multilabel sewer defect classification. In this study, we employ the kernel patch attention optical flow estimator~\cite{luo2022learning} to efficiently create a motion image dataset. Subsequently, we train our DS-MSHViT network using both motion and RGB datasets, aiming to enhance the robustness of sewer defect classification.

\section{Method}\label{Section:Proposed}
\begin{figure*}
    \centering
    \includegraphics[width=\textwidth]{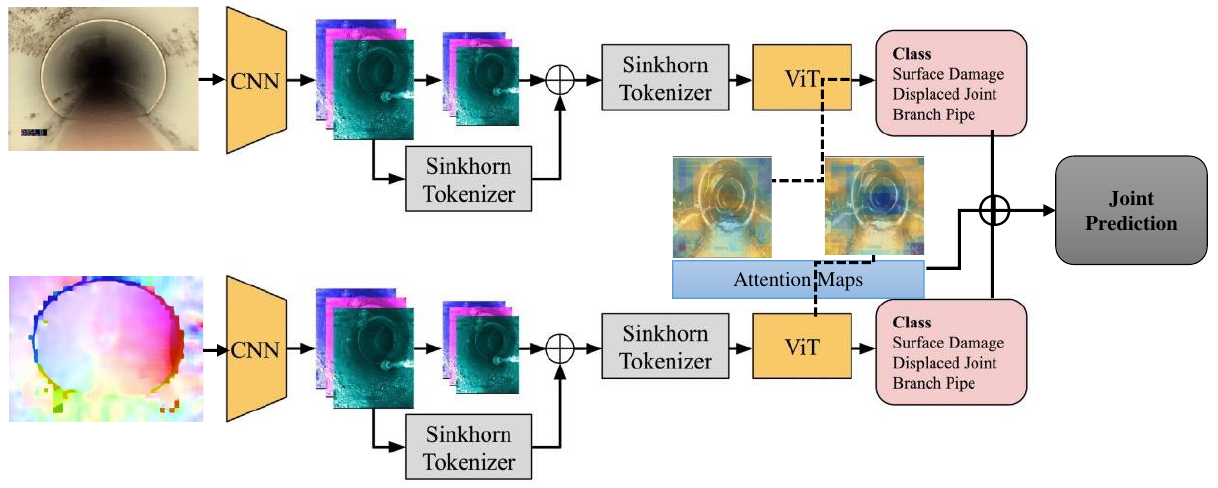}
    \caption{\textbf{Proposed DS-MSHViT Architecture}: The proposed architecture introduces Dual-Stream Attention Transformers, which connect two Multi-Scale Hybrid Vision Transformer models~\cite{haurum2022multi} in parallel to process an RGB frame and a motion frame separately at high resolution. By utilizing both the spatial details from the RGB input and temporal cues from the motion input, this dual-stream training aims to improve the RGB network classification performance. }
    \label{fig:architecture}
\end{figure*}

Fig.~\ref{fig:architecture} illustrates the proposed dual-stream attention transformers architecture. The proposed method uses two identical neural network models to process the RGB frames and motion frames separately. For each input frame, the respective network generates an attention map that highlights the regions of the image most relevant to the classification task. The key insight is that the attention maps for the RGB and motion streams should focus on similar areas since they represent the same underlying content. In the following subsections, we explain the modules of the dual-stream architecture in detail.

\subsection{CNNs for Extracting Image-specific Inductive Biases}
In the field of automated sewer inspection, convolutional neural networks (CNNs) have emerged as the dominant method for analysis of sewer image data.
A key component that makes CNNs effective for this domain is the convolutional layers. Convolutional layers capture local spatial semantics and patterns within images through the use of convolutional filters.

Given an input image $\mathbf{x}$ of size $C \times H \times W$ (with C channels, height H and width W) from a training dataset, CNNs are designed to learn relevant relationships between features in the image.
CNNs encode certain inductive biases through their architecture that allow them to learn these relationships effectively for images:

\begin{itemize}
\item Locality is built into CNNs through the convolution operation. This operates by sliding a small $p \times p$ filter kernel over the input image with a stride of $p$ pixels. The output is a feature map $F$ where each value depends only on a local $p \times p$ region of the input image $\mathbf{x}$.
\item Weight sharing is used by employing the same filter weights across the entire input image. This greatly reduces the number of learnable parameters to generate the feature map $F$.
\item Pooling layers (e.g. max pooling) provide spatial invariance by summarizing the activations over local regions of the feature map $F$. This allows the CNN to be robust to small translations in $\mathbf{x}$.
\item CNNs are organized into hierarchical layers, with each layer learning increasingly abstract representations of the input image in its feature maps. Early layers focus on low-level features like edges, while deeper layers combine these to detect higher-level concepts like objects. This hierarchical structure allows the CNN to learn highly complex relationships between image features.
\end{itemize}

\subsection{ViT for Extracting Non-local Spatial Semantics}
Unlike convolutional neural networks (CNNs) which leverage local spatial relationships through convolutions, Vision Transformers (ViTs) do not contain any convolutional layers. Instead, a ViT learns directly from the global context of image patches through self-attention mechanisms.

Given an input image $\mathbf{x} \in \mathbb{R}^{C \times H \times W}$, it is first split into small non-overlapping patches of size $P\times P$, resulting in $\frac{HW}{P^2}$ total patches. Each $P \times P \times C$ patch is then flattened into a 1D vector $T \in \mathbb{R}^{D}$ where $D=P^2C$ is the patch dimension.
These patch vectors ${T}$ are mapped to $N$ embedded token vectors ${T_p} \in \mathbb{R}^{D\times N}$ by applying a trainable linear projection to each. This embedding represents each patch independently.
In addition, a special classification token $x_{cls} \in \mathbb{R}^D$ is added to the sequence of embedded patch tokens. This token is used to represent the entire image, and its output will be passed to the classification head.
To retain spatial information, a learnable positional encoding $E_{pos} \in \mathbb{R}^{D \times N+1}$ is added to the patch and class tokens. This encoding assigns each token a unique position-specific representation. The final token representation followed by positional encoding can be expressed as: 
\begin{equation}
    Z_0 = [x_{cls} \oplus T_p] + E_{pos}, 
\end{equation}
where $\oplus$ represents concatenation.

The Transformer encoder in ViT applies multi-head self-attention on the input sequence of patch, class, and position-encoded tokens. With multi-head attention, the model can jointly attend to different representation subspaces at different positions, increasing its expressive power.
Specifically, the Transformer encoder contains $L$ layers with each layer having two sub-layers. The first is a multi-head self-attention layer that aggregates information from the full sequence of input tokens. The second sub-layer is a position-wise feedforward network that further processes the attended features.
The output of the Transformer encoder contains global contextual representations of the entire image, which can be used by the classification head to predict the image class. By modeling long-range dependencies between patches, the ViT can recognize image features regardless of their spatial locations.

\subsection{MSHViT for Multi-label Sewer Defect Classification}

Multi-Scale Hybrid Vision Transformer (MSHViT) architecture fuses CNN and ViT models along with clustering-based tokenization for multi-label sewer defect classification. MSHViT appends ViT modules at different stages of a CNN backbone to enable feature aggregation across scales as well as introduces a Sinkhorn tokenizer based on clustering to replace the standard patch tokenizer in ViTs.

The Sinkhorn tokenizer reduces redundant tokens generated from the CNN features before feeding them into the ViT layers by computing cosine similarity scores $V$ between $N$ patch tokens $T_p$ and $K$ cluster centers $C\in \mathbb{R}^{D \times K}$. Next, the soft assignment matrix $Q\in \mathbb{R}_{+}^{K \times N}$ is computed based on $V$.
Finally, employing Sinkhorn distances to softly assign the input tokens $T_p$ to a learned set of cluster centers $C$, resulting in a smaller condensed $K$ new tokens $T_s$ as:
\begin{equation}
    T_s = T_p Q^{\intercal}.
\end{equation}
This tokenized feature representation retains non-local semantics while removing unnecessary information. 

MSHViT takes advantage of natural scale pyramids within CNN backbones by selecting $i$-th layer feature map $F_i$ just before the pooling operation. Different layers of feature maps naturally represent different scales based on how much downsampling has occurred. MSHViT utilizes cross-scale connections to propagate features across the ViT layers in a top-down pathway. This allows hierarchical sharing of information between scales at three different stages, i.e., linearly embedded CNN features $T_p$, the Sinkhorn token $T_s$, or the final token embedding $Z_L$ at depth $L$.

During training, the MSHViT minimizes a binary cross-entropy (BCE) loss which penalizes predictions that diverge from the true label. Formally, BCE can be written for a single input-label pair $(\mathbf{x}, y)$ as:
\begin{equation}
\mathcal{L}(\mathbf{x}, y) = - [ y \log(\hat{y}) + (1 - y) \log(1 - \hat{y}) ],
\end{equation}
where 
$\mathbf{x}$ is the input image,
$y$ is the ground truth label,
$\hat{y}$ is the predicted probability of a class from the model for input $\mathbf{x}$.
By taking the negative log-likelihood, the loss is minimized when the prediction matches the true label.

\subsection{Generating Motion Images with Kernel Patch Attention}

Optical flow estimates dense motion fields between video frames. Given consecutive frames $\mathbf{x}_{t-1}$ and $\mathbf{x_t}$, it outputs a per-pixel flow field that encodes displacements between corresponding pixels. This compactly represents apparent motion from $\mathbf{x}_{t-1}$ to $\mathbf{x_t}$ as an image.

Recent work, Kernel Patch Attention (KPA), improves optical flow estimation by explicitly modeling spatial relations and motion affinities. The key idea is to apply attention within local patches to capture region-based context and use that to refine motion features. The fused motion feature $\hat{\mathbf{f}}_m$ is obtained by applying the kernel-predicting attention function $\mathcal{F}_{KPA}$ to the context feature $\mathbf{f}_c$ and original motion feature $\mathbf{f}_m$. This fuses information from both sources into an enhanced motion representation.

Specifically, the KPA extracts multi-scale features from an input image pair ($\mathbf{x}_{t-1}, \mathbf{x_t}$), then use two residual-based encoders to extract a feature pair $(\mathbf{f}_1, \mathbf{f}_2)$ and context feature $\mathbf{f}_c$. Next, 4D correlation volumes on all vector pairs within the multi-scale features $\mathbf{f}_1$ and $\mathbf{f}_2$. 
This learns to match pixels between the images in a multi-scale manner. Finally, the refined motion features integrating spatial constraints are then fed to a decoder to estimate residual flows iteratively.

To utilize motion information for training neural networks, the fused motion feature $\hat{\mathbf{f}}_m$ is converted into an RGB optical flow image $\mathbf{x}_{o}$. This is done by reshaping $\hat{\mathbf{f}}_m$ into $H\times W \times 2$ tensors containing displacement vectors $(dx, dy)$ for each pixel. The magnitude $M$ and orientation $\theta$ of these vectors are computed and encoded into the $R$ and $G$ channels as $R=M \cos(\theta)$ and $G = M sin(\theta)$. The $B$ channel is set to 1 for visualization. Finally, the $[R,G,B]$ values are rescaled to $[0,255]$ to obtain a color-coded motion image. This compactly represents the dense predicted motion field, with color indicating direction and intensity representing the speed of each motion vector.

Note that it is also possible to generate motion images from synchronous LiDAR streams. In this case, the LiDAR streams would first need to be converted to depth images, as shown in~\cite{islam2021pedestrian}. Optical flow methods could then be applied to the depth images to generate corresponding motion representations.

\subsection{DS-MSHViT for Multi-label Sewer Defect Classification}


Sewer defect datasets collected by CCTV robots are often low resolution and affected by motion blur. To address this, our proposed Dual Stream MSHViT (DS-MSHViT) incorporates two input streams \textendash RGB and optical flow motion images,  to learn robust attention maps for defect classification.
We generate two training datasets \textendash RGB dataset $\mathcal{D}_r$ and optical flow dataset $\mathcal{D}_o$ by running a pretrained KPA model on the videos.
DS-MSHViT offers a dual training mechanism that takes an RGB image $\mathbf{x}_r \in \mathcal{D}_r$ and a corresponding motion image $\mathbf{x}_o \in \mathcal{D}_o$ as inputs.
The two streams allow the model to leverage complementary information from both modalities to handle the challenges of low resolution and motion blur. 

A straightforward approach to utilize the two modalities would be to independently train separate MSHViT networks and ensemble their predictions. However, we argue that the key limitation of MSHViT is generating inaccurate attention maps from the self-attention layers. Furthermore, motion images lack rich features, resulting in poor classification accuracy alone. Naively combining with a motion-trained MSHViT can introduce many false positives, degrading classification performance.

To address this, DS-MSHViT focuses on improving the robustness of attention maps using both modalities. Rather than late fusion, we introduce an early attention alignment mechanism. This novel loss function optimizes the consistency between the RGB and motion attention maps. By encouraging agreement between the complementary streams during training, DS-MSHViT produces more accurate and consistent attention, leading to improved classification, especially for challenging sequential images.
Given an input pair ($\mathbf{x}_r$, $\mathbf{x}_o$), along with the ground truth label $y$, the total loss $\mathcal{L}$ is defined as:
\begin{equation}
    \mathcal{L}(\mathbf{x}_r, \mathbf{x}_o, y) = \mathcal{L}_r(\hat{y}_r, y) + \mathcal{L}_o(\hat{y}_o, y) + \mathcal{L}_a(F_r^i, F_o^i),
\end{equation}
where $\mathcal{L}_r$ is the classification loss between the RGB model's prediction $\hat{y}_r$ and the label $y$, $\mathcal{L}_o$ is the classification loss between the motion model's prediction $\hat{y}_o$ and $y$, and $\mathcal{L}_a$ is the attention alignment loss between the RGB attention map ${F}^i_r$ and the motion attention map ${F}^i_o$.
By minimizing both the classification losses $\mathcal{L}_r$ and $\mathcal{L}_o$, the model is trained to predict the correct labels $y$ from both modalities. Additionally, the attention alignment loss $\mathcal{L}_a$ minimizes $i$-th layer the feature distance between $F_r^i$ and $F_o^i$, encouraging the attention maps to focus on consistent salient regions. This joint optimization framework enables robust fused representations by aligning the complementary RGB and motion streams.

In addition, the overall loss function includes a standard cross-entropy classification loss that penalizes discrepancies between the predicted class probabilities and the ground truth labels. Optimizing both the attention alignment and classification losses jointly enables the model to learn effective fused representations that integrate information from both the spatial RGB and temporal motion streams.
By explicitly training the model to align the modalities through attention map consistency, the aim is to achieve better feature learning and improved classification accuracy compared to approaches that do not align the dual-stream attention. The joint optimization framework allows the model to learn in a way that leverages the complementary strengths of the RGB and motion inputs.
\section{Experiments}\label{Section:Experiment}
The experiments and benchmarking were performed using a server with an AMD EPYC 7713 64-Core CPU, 1TB of RAM, and an NVIDIA A100 GPU with 80GB of memory. The code for DS-MSHViT is available at~\url{https://github.com/RedwanNewaz/ds_mshvit}. We are unable to share the Carencro dataset directly due to a confidentiality agreement, but interested researchers can contact the Sewerage and Water Board of New Orleans to request access.

\subsection{Training Losses}
We evaluate our proposed DS-MSHViT model by applying it to multiple MSHViT backbones with different CNN architectures.
Each DS-MSHVIT model is trained for 40 epochs, which is equivalent to around 160,000 training steps. The total loss is calculated based on three components:
\begin{itemize}
    \item \textbf{RGB network loss:} This loss measures the difference between the predicted label on the RGB stream and the ground truth label,
    \item \textbf{Optical network loss:} This loss measures the difference between the predicted label on the optical flow stream and the ground truth label,
    \item \textbf{Attention map loss:} This loss measures the difference between the attention maps generated by the RGB and optical networks.
\end{itemize}

\begin{figure}[h]
  \centering
  \begin{tikzpicture}
    \begin{axis}[
      xlabel={Steps},
      ylabel={Loss},
      legend entries={Attention loss, Optical loss, RGB loss, Total loss},
      legend pos=north east,
    ]
      
      \addplot table [x=step, y=attention_loss, col sep=comma] {Figs/plot_data/combined_losses.csv};
      \addplot table [x=step, y=optical_loss, col sep=comma] {Figs/plot_data/combined_losses.csv};
      \addplot table [x=step, y=rgb_loss, col sep=comma] {Figs/plot_data/combined_losses.csv};
      \addplot table [x=step, y=total_loss, col sep=comma] {Figs/plot_data/combined_losses.csv};

    \end{axis}
  \end{tikzpicture}
  \caption{\textbf{Training Loss Analysis for Proposed DS-MSHViT Architecture}: Training losses are calculated over 40 epochs and around 160K training steps. The total loss is calculated based on three losses, i.e., RGB network loss, Optical network loss, and Attention map loss between RGB and Optical networks. The training losses for all three objectives exhibit a downward trend, signifying that the model progressively improves on the task. The reductions in the losses imply that the network learns increasingly discriminative features and alignment that better optimize the objectives.
  }\label{fig:train_loss}
\end{figure}
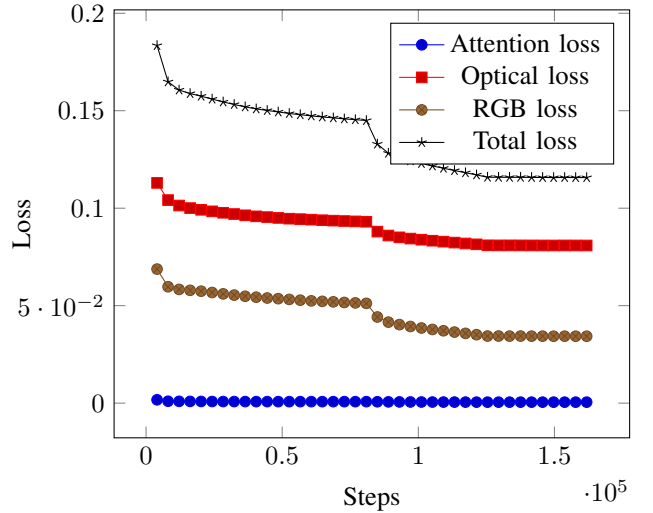

Fig.~\ref{fig:train_loss} shows training loss curves for our proposed DS-MSHVIT with a ResNet-50 backbone.
Over the 40 epochs of training Attention map loss reduces from $1.71\times 10^{-03}$ to $5.00\times 10^{-04}$,  RGB network loss reduces from $1.71\times 10^{-03}$ to $5.00\times 10^{-04}$, Optical network loss reduces from $1.13\times 10^{-01}$ to $8.09\times 10^{-02}$, and Total loss reduces from $6.88\times 10^{-02}$ to $3.43\times 10^{-02}$.
We can see that all three losses decrease significantly over the course of training, indicating that the model is learning to perform its classification task more accurately.

\subsection{Performance Evaluation on the Sewer-ML Dataset}
We thoroughly benchmark our model performance on existing MSHViT model with several commonly used backbone architectures in the image classification literature such as ResNet, TResNet, the HViTlike models BoTNet-50-S1 and CoAtNet-\{0, 1\}, as well as the original HViT structure. We also compare the previously published results on MSHViT with patch based tokenizer and the Sinkhorn tokenizer.
Based on table~\ref{tab:benchmark}, the DS-MSHVIT model achieves the best performance overall across both validation and test sets on the Sewer-ML datasets. Full test results are only available through the ended Sewer-ML competition, we therefore represent missing test scores with ``-" in table~\ref{tab:benchmark}.

On the SewerML validation set, DS-MSHVIT with BoTNet-50-S1, ResNet-50, and ResNet-101 backbones achieves F2-CIW scores of 62.15, 62.12, and 63.15 respectively, along with F1-Normal scores of 92.96, 92.96, and 93.08. In comparison, baseline BoTNet-50-S1 and MSHViT with ResNet-101, ResNet-50 backbones attain lower F2-CIW scores of 61.62, 61.25, 61.68, and lower F1-Normal scores of 92.92, 92.50, 92.44. This highlights the consistent improvements from the proposed DS-MSHVIT architecture across backbone networks.
This represents an improvement of 0.53\% to 1.53\% in F2-CIW and 0.04\% to 0.58\% in F1-Normal metrics.

On the SewerML test set, we observe a similar trend: DS-MSHVIT with ResNet-101 backbone sees the largest improvements, i,e., 62.50 F2-CIW score 92.51 F1-Normal score.
This represents an improvement of 2.57\% in F2-CIW score and 0.32\% in F1-Normal score compared to MSHViT with the ResNet-101 backbone.
Therefore, we can conclude that the dual-stream architecture allows DS-MSHVIT to learn more complex relationships between the RGB and motion images, which results in improved performance on the Sewer-ML dataset.

\begin{table}[h]
\centering
\begin{adjustbox}{max width=0.5\textwidth}
\begin{tabular}{@{}lcccc@{}}
\toprule
\textbf{Model}          & \multicolumn{1}{c}{\textbf{\begin{tabular}[c]{@{}c@{}}Val\\ F2-CIW\end{tabular}}} & \multicolumn{1}{c}{\textbf{\begin{tabular}[c]{@{}c@{}}Val\\ F1-Normal\end{tabular}}} & \multicolumn{1}{c}{\textbf{\begin{tabular}[c]{@{}c@{}}Test\\ F2-CIW\end{tabular}}} & \multicolumn{1}{c}{\textbf{\begin{tabular}[c]{@{}c@{}}Test\\ F1-Normal\end{tabular}}} \\ \midrule

Benchmark               & 55.36              & 91.32                  & 55.11               & 90.94                   \\
CT-GAT*                 & 61.70               & 91.94                  & 60.27               & 91.61                   \\
ResNet-50-HViT-Patch    & 59.87              & 92.41                  & 57.58               & 91.99                   \\
ResNet-50-HViT-Sinkhorn & 60.42              & 92.41                  & 58.74               & 92.07                   \\
BotNet-50-S1            & 61.62              & 92.92                  & 59.69               & 92.49          \\
CoAtNet-0               & 57.82              & 92.28                  & 56.53               & 91.94                   \\
CoAtNet-1               & 59.37              & 92.50                   & 57.42               & 91.11                   \\
ResNet-18               & 58.60               & 92.34                  & 56.62               & 91.88                   \\
ResNet-18 + MSHVIT             & 59.87              & 92.42                  & 58.18               & 92.12                   \\
ResNet-34               & 60.98              & 92.72                  & 59.18               & 92.30                    \\
ResNet-34 + MSHVIT             & 61.65              & 92.76                  & 59.91               & 92.03                   \\
ResNet-50               & 59.28              & 92.44                  & 57.58               & 92.11                   \\
ResNet-50 + MSHVIT             & 61.68              & 92.44                  & 60.11               & 92.13                   \\
ResNet-101              & 60.06              & 92.48                  & 58.01               & 92.13                   \\
ResNet-101 + MSHVIT             & 61.25              & 92.50                   & 59.93               & 92.19                   \\
TResNet-M               & 58.04              & 92.22                  & 56.08               & 91.90                    \\
TResNet-M + MSHVIT             & 58.68              & 92.25                  & 56.93               & 91.84                   \\
TResNet-L               & 59.17              & 92.36                  & 56.97               & 92.00                      \\
TResNet-L + MSHVIT             & 59.19              & 92.27                  & 57.16               & 91.87  
\\
CoAtNet-0 + DS-MSHVIT      & 58.77     & 92.51        & \textbf{-}      & -                   \\ 
TResNet-L + DS-MSHVIT    & 59.18     & 92.40        & \textbf{-}      & -                   \\ 
TResNet-M + DS-MSHVIT    & 59.07     & 92.37        & \textbf{-}      & -                   \\ 
ResNet-18 + DS-MSHVIT    & 58.37     & 92.24        & \textbf{-}      & -                   \\ 
ResNet-34 + DS-MSHVIT    & 58.19     & 92.35        & \textbf{-}      & -                   \\ 
\textbf{BotNet-50-S1 + DS-MSHVIT}      & \textbf{62.15}     & \textbf{92.86}         & 60.22      & 92.37                   \\ 
\textbf{ResNet-50 + DS-MSHVIT}      & \textbf{62.12}     & \textbf{92.96}         & \textbf{60.47}      & 92.41                   \\ 
\textbf{ResNet-101 + DS-MSHVIT}      & \textbf{63.15}     & \textbf{93.08}         & \textbf{62.50}     &    \textbf{92.51}               \\ \bottomrule
\end{tabular}
\end{adjustbox}
\caption{DS-MSHVIT sets a new state-of-the-art on this sewer defect classification dataset, significantly outperforming prior benchmarks, especially on the more challenging defective sample identification. It combines strong performance on both defect and non-defect categories across validation and test sets.}\label{tab:benchmark}
\end{table}

\subsection{Cross Dataset Evaluation}
To assess the generalization ability of models pre-trained on the Sewer-ML dataset, we generated a new dataset comprising 50,000 images from inspection videos in the city of Carencro, LA, USA. This dataset includes multi-label annotations for five defect classes: deformation (DE), displaced joint (FS), settled deposits (AF), branch pipe (GR), and connection with construction changes (OK). We use this dataset to evaluate the impact of pre-training on large-scale datasets for cross-dataset performance.
Since we did not skip any frames during annotation, this dataset represents realistic conditions where defects must be identified at real-time speeds. The consecutive frames provide a robust test of model generalization under domain shift from the original Sewer-ML data.

Table~\ref{tab:cross_data} compares the performance of various pre-trained models on the cross-dataset, focusing on mAP (Mean Average Precision), average precision, and average recall as evaluation metrics. The aim is to assess the impact of dual-stream training on model performance in these metrics. The DS-MSHVIT model stands out as the top performer with the highest mAP of 32.98\%. These results demonstrate that the generalizability of DS-MSHViT is much better than the baseline MSHViT. The DS-MSHViT model can be readily adapted to new domains with little to no additional training, while the MSHViT lacks robustness under domain shift. 
\begin{table*}[h]
\begin{adjustbox}{max width=\textwidth}
\begin{tabular}{@{}lcccccccccc@{}}
\toprule
\multicolumn{1}{c}{\textbf{Model}} & \multicolumn{2}{c}{\textbf{F1-Normal}}                    & \multicolumn{2}{c}{\textbf{F2-Normal}}                    & \multicolumn{2}{c}{\textbf{mAP}}                           & \multicolumn{2}{c}{\textbf{OP}}                            & \multicolumn{2}{c}{\textbf{OR}}                            \\ \midrule
                                   & \multicolumn{1}{l}{MSHViT} & \multicolumn{1}{l}{DS-MSHViT} & \multicolumn{1}{l}{MSHViT} & \multicolumn{1}{l}{DS-MSHViT} & \multicolumn{1}{l}{MSHViT} & \multicolumn{1}{l}{DS-MSHViT} & \multicolumn{1}{l}{MSHViT} & \multicolumn{1}{l}{DS-MSHViT} & \multicolumn{1}{l}{MSHViT} & \multicolumn{1}{l}{DS-MSHViT} \\
BotNet-50-S1                       & 88.925                     & \textbf{89.651}               & 95.255                     & \textbf{95.315}               & \textbf{29.512}            & 27.959                        & 80.116                     & \textbf{81.368}               & 76.172                     & \textbf{77.366}               \\
CoAtNet-0                          & \textbf{88.907}            & 88.758                        & 95.135                     & \textbf{95.174}               & \textbf{26.962}            & 15.046                        & \textbf{80.136}            & 79.790                         & \textbf{76.190}             & 75.862                        \\
ResNet-18                          & \textbf{89.122}            & 89.068                        & 94.799                     & \textbf{95.253}               & 26.361                     & \textbf{26.784}               & 80.363                     & \textbf{80.407}               & 76.422                     & \textbf{76.454}               \\
ResNet-34                          & \textbf{89.320}            & 88.734                        & \textbf{95.246}            & 94.624                        & \textbf{29.461}            & 17.383                        & \textbf{80.674}            & 79.613                        & \textbf{76.717}            & 75.693                        \\
ResNet-50                          & 88.890                     & \textbf{89.388}               & 95.088                     & \textbf{95.463}               & 22.938                     & \textbf{32.977}               & 80.033                     & \textbf{81.039}               & 76.094                     & \textbf{77.049}               \\
ResNet-101                         & 88.951                     & \textbf{90.139}               & 95.255                     & \textbf{95.375}               & 24.498                     & \textbf{33.098}               & 80.145                     & \textbf{81.945}               & 76.199                     & \textbf{77.920}               \\
TResNet-M                          & 89.522                     & \textbf{90.063}               & 95.095                     & \textbf{95.255}               & \textbf{27.727}            & 23.012                        & 80.892                     & \textbf{81.186}               & 76.919                     & \textbf{77.206}               \\
TResNet-L                          & \textbf{89.375}            & 88.262                        & \textbf{95.421}            & 93.608                        & \textbf{31.313}            & 28.444                        & \textbf{80.871}            & 79.210                         & \textbf{76.888}            & 75.315                        \\ \bottomrule
\end{tabular}
\end{adjustbox}
\caption{This table shows the results of evaluating different models on a cross-dataset, highlighting the benefits of the dual-stream setup with MSHVIT. The proposed DS-MSHVIT with ResNet-101 backbone model achieves the best F1-Normal score of 90.139\% and the best mAP of 33.098\%. The proposed DS-MSHVIT with ResNet-50 backbone model achieves the best F2-Normal score of 95.463\%. 
DS-MSHVIT with ResNet-101 backbone achieves the highest overall precision of 81.945\% as well as the highest overall recall of 77.920\%.
Overall, DS-MSHVIT strikes the best balance of accuracy and recall across this cross-dataset evaluation.
}\label{tab:cross_data}
\end{table*}
\subsection{Ablation Studies}
We conducted a series of ablation studies to determine the role of individual networks in predicting sewer deficiencies. We also examine which is the best layer where attention should be regularized in the DS-MSHViT architecture. All tests are conducted on the Sewer-ML validation set using a ResNet-50 backbone.
\paragraph{Finding the best network branch for classification task}
The dual-stream setup consists of two separate streams: an RGB stream and a motion stream. The RGB stream processes RGB images and the motion stream processes motion information, such as optical flow.
Once networks are fully trained the final prediction can be generated in three ways: 
(i) using only the RGB branch outputs,
(ii) using only the optical flow branch outputs, or
(iii) fusing the outputs from both network branches.
This can be formally expressed as: 
\begin{equation}
    \hat{y} = \alpha \hat{y}_r + (1 - \alpha) \hat{y}_o,
\end{equation}
where $\hat{y}_r$ and $\hat{y}_o$ are predicted scores from RGB and motion stream networks. 
The hyperparameter $\alpha$ determines the ratio of pairing between the RGB and motion streams during training. When $\alpha=0$, the RGB and motion streams are fully paired, which means that each RGB image is paired with a motion image. When $\alpha=1$, the RGB and motion streams are not paired at all, which means that each RGB image is processed independently. We can also represent $\alpha$ as a learning parameter and can be updated using the sub-gradient method during training.   
\begin{table}[h]
\centering
\begin{adjustbox}{max width=0.5\textwidth}
\begin{tabular}{@{}llccc@{}}
\toprule
\textbf{First Network} & \textbf{Second Network} & \multicolumn{1}{c}{$\alpha$} & \textbf{F1-Normal} & \textbf{F2-CIW} \\ \midrule
RGB Image (t)       & RGB Image (t+1)      & 0                                     & 83.01              & 41.56           \\
RGB Image (t)       & RGB Image (t+1)      & Learned: 0.4956                       & 90.41              & 49.47           \\
RGB Image (t)       & RGB Image (t+1)      & 1                                     & 92.68              & 60.79           \\
RGB Image (t)       & Motion Image (t+1)     & 0                                   & 81.49              & 20.75           \\
RGB Image (t)       & Motion Image (t+1)     & 0.5                                   & 90.75              & 41.04           \\
RGB Image (t)       & Motion Image (t+1)     & Learned: 0.8194                       & 92.90              & 60.92           \\
RGB Image (t)       & Motion Image (t+1)     & 1                                     & \textbf{92.96 }             & \textbf{62.12 }          \\ \bottomrule
\end{tabular}
\end{adjustbox}
\caption{In the proposed dual-stream setup, a Resnet-50 backbone with MSHVIT networks is utilized. The parameter $\alpha$ determines the ratio of pairing between the RGB and motion streams during training. $\alpha$ is learned and validated on the cross-validation set. The results in the table show that setting $\alpha=1$, which corresponds to fully pairing the RGB and motion streams, achieves the best performance. Therefore, the RGB and motion streams are fully paired during training, but only the RGB stream is used for testing. 
}\label{tab:alpha}

\end{table}

\noindent Table~\ref{tab:alpha} answers two key questions: 
\begin{itemize}
    \item \textit{How does pairing two identical RGB streams in the DS-MSHViT architecture compare to the proposed RGB-motion pairing? }
    \item \textit{Can inference fusion between the RGB and motion branches outperform using only the RGB branch? }
\end{itemize}
The results in table~\ref{tab:alpha} show that setting $\alpha=1$ with the proposed RGB-motion pairing achieves the best F1-Normal score of 92.96\% and the best F2-CIW score 62.15\%. 
This also indicates that pairing the RGB and motion streams during training, while using only the RGB stream for testing, is an optimal strategy.
Analyzing the learned $\alpha$ reveals that pairing identical RGB streams causes competition between branches. With RGB-motion, the motion branch acts as a regularizer for the RGB attention maps to focus on discriminative areas.
This regularization effect during training enables the RGB stream to generalize better on its own during testing, without needing the motion stream. Thus, the dual-stream approach exploits the complementary information from motion to train a better RGB model, while retaining the efficiency of a single-stream architecture, i.e., the RGB network branch, at test time. 

\paragraph{Finding the best attention layer}

\begin{figure}
 \centering
    \begin{tikzpicture}
        \begin{axis}[
            ybar,
            bar width=0.4cm,
            xlabel={\textbf{Attention Layers}},
            ylabel={Scores},
            symbolic x coords={Layer 3 + Layer 4, Layer 4, Layer 3},
            xtick=data,
            nodes near coords,
            nodes near coords align={vertical},
            legend style={at={(0.1,0.3)},anchor=south west}
            ]
            \addplot coordinates {(Layer 3 + Layer 4, 92.75) (Layer 4, 92.68) (Layer 3, 92.96)};
            \addplot coordinates {(Layer 3 + Layer 4, 61.63) (Layer 4, 61.00) (Layer 3, 62.12)};
            \legend{F1-Normal, F2-CIW}
        \end{axis}
    \end{tikzpicture}
    \caption{\textbf{F1-Normal and F2-CIW Scores for Different Attention Layers}: Proposed DS-MSHVIT model chooses to minimize the attention loss at Layer 3 only. This provides the best trade-off \textemdash competitive F1-Normal score while substantially improving F2-CIW score on challenging Sewer-ML validation dataset.}
    \label{tab:attn}
\end{figure}
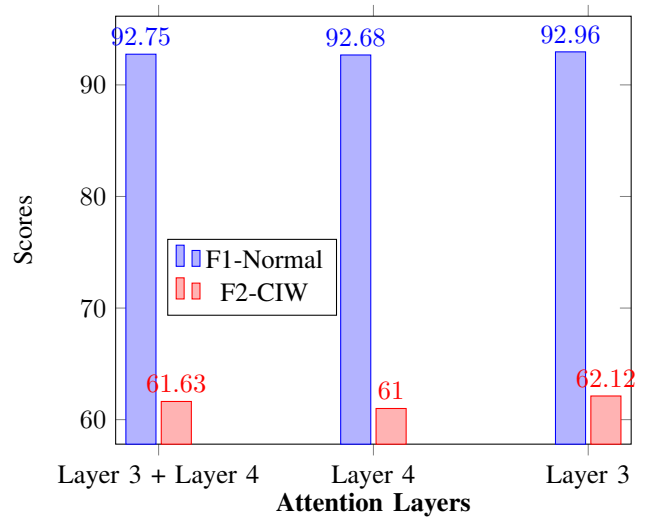

In Fig.~\ref{tab:attn}, we present a comparison of F1 and F2-CIW scores on the Sewer-ML validation dataset, focusing on the minimization of attention loss at different layers within the DS-MSHVIT network. The key findings are as follows:
Achieving the highest F1-Normal score of 92.75\% on the cross-validation dataset is possible when minimizing attention loss at both Layer 3 and Layer 4. However, when it comes to the best F2-CIW score, which reaches 62.12\%, it is attained by minimizing attention loss at Layer 3 exclusively. On the other hand, minimizing attention loss at Layer 4 alone results in the poorest F2-CIW score performance. This underscores the importance of Layer 3 attention for achieving generalization across datasets.


\begin{figure*}
    \centering
    \begin{subfigure}{.3\textwidth}
        \centering
        \includegraphics[width=\linewidth]{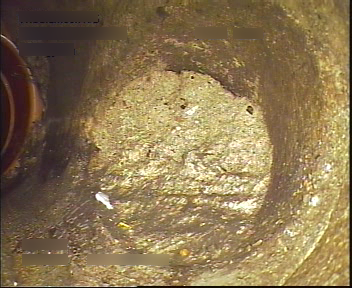}
        \label{fig:subfig_a}
        \begin{adjustbox}{max height=1cm}
        \begin{tabular}{|c|c|c|}
            \hline
            \textbf{GT} & \textbf{MSHViT} & \textbf{DS-MSHViT} \\ \midrule
            RB          & -               & -                  \\
            OB          & -               & OB                 \\
            FS          & -               & FS                 \\
            GR          & -               & GR                 \\
            OK          & -               & OK                 \\ \bottomrule
            \hline
        \end{tabular}
        \end{adjustbox}
    \end{subfigure}
    \begin{subfigure}{.3\textwidth}
        \centering
        \includegraphics[width=\linewidth]{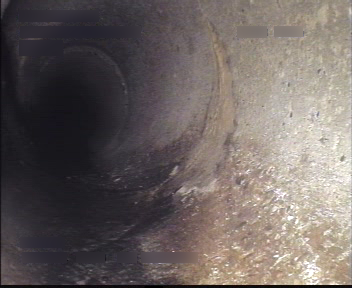}
        
        \begin{adjustbox}{max height=1cm}
        \begin{tabular}{|c|c|c|}
            \hline
            \textbf{GT} & \textbf{MSHViT} & \textbf{DS-MSHViT} \\ \midrule
            RB          & -               & RB                  \\
            FS          & -               & FS                 \\
            BE          & -               & BE                 \\
            GR          & GR               & GR                 \\
                       &                 &                  \\ \bottomrule
            \hline
        \end{tabular}
        \end{adjustbox}
        \label{fig:subfig_b}
    \end{subfigure}
    \begin{subfigure}{.3\textwidth}
        \centering
        \includegraphics[width=\linewidth]{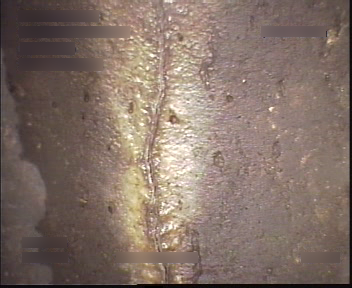}
        \begin{adjustbox}{max height=1cm}
        \begin{tabular}{|c|c|c|}
            \hline
            \textbf{GT} & \textbf{MSHViT} & \textbf{DS-MSHViT} \\ \midrule
            RB          & -               & RB                  \\
            OB          & -               & OB                 \\
            IN          & -               & -                 \\
            BE          & -               & BE                 \\
                       &                 &                   \\ \bottomrule
            \hline
        \end{tabular}
        \end{adjustbox}
        \label{fig:subfig_c}
    \end{subfigure}
    
    \vspace{1em} 
    \begin{subfigure}{.3\textwidth}
        \centering
        \includegraphics[width=\linewidth]{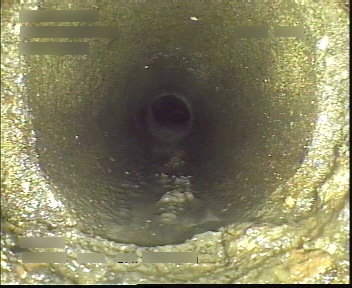}
        \begin{adjustbox}{max height=1cm}
        \begin{tabular}{|c|c|c|}
            \hline
            \textbf{GT} & \textbf{MSHViT} & \textbf{DS-MSHViT} \\ \midrule
            OB          & -               & OB                 \\
            FS          & -               & FS                 \\
            PH          & -               & PH                 \\
                       &                &                   \\ \bottomrule
            \hline
        \end{tabular}
        \end{adjustbox}
        \label{fig:subfig_d}
    \end{subfigure}
    \begin{subfigure}{.3\textwidth}
        \centering
        \includegraphics[width=\linewidth]{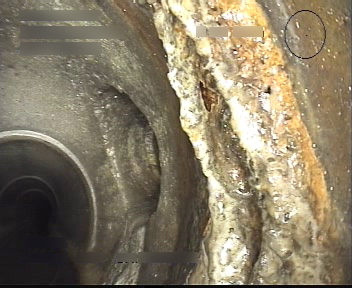}
        \begin{adjustbox}{max height=1cm}
        \begin{tabular}{|c|c|c|}
            \hline
            \textbf{GT} & \textbf{MSHViT} & \textbf{DS-MSHViT} \\ \midrule
            GR          & -               & -                 \\
            FS          & FS               & FS                 \\
            IN          & -               & IN                 \\
            BE          & -               & BE                 \\ \bottomrule
            \hline
        \end{tabular}
        \end{adjustbox}
        \label{fig:subfig_e}
    \end{subfigure}
    \begin{subfigure}{.3\textwidth}
        \centering
        \includegraphics[width=\linewidth]{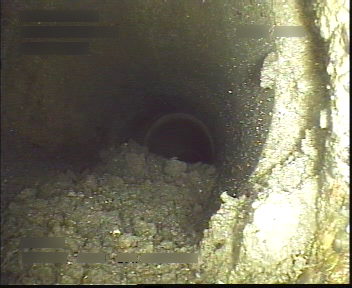}
        \begin{adjustbox}{max height=1cm}
        \begin{tabular}{|c|c|c|}
            \hline
            \textbf{GT} & \textbf{MSHViT} & \textbf{DS-MSHViT} \\ \midrule
            OB          & -               & -                  \\
            FS          & -               & FS                 \\
            PH          & -               & PH                 \\
                       &                &                  \\ \bottomrule
            \hline
        \end{tabular}
        \end{adjustbox}
        \label{fig:subfig_f}
    \end{subfigure}
    \caption{\textbf{Visual Analysis of DS-MSHViT and MSHViT}: Prediction outcomes on sewer image validation data comparing DS-MSHViT and MSHViT models with ResNet-101 backbones. 
    The first column of each table represents the ground truth (GT) annotation.
    The subfigures clearly demonstrate DS-MSHViT's improved multi-label classification performance, with significantly more successful predictions and fewer failure cases than MSHViT. Defect type codes follow the original notation defined in~\cite{haurum2022multi}.}
    \label{fig:qual-example}
\end{figure*}

Given our objective of enhancing generalization, the DS-MSHVIT model opts to minimize attention loss at Layer 3 exclusively. This approach strikes the optimal balance: delivering a competitive F1-Normal score while significantly improving the challenging F2-CIW score on the Sewer-ML validation dataset.
Through the selective regularization of Layer 3 attention, DS-MSHVIT succeeds in learning representations that exhibit increased invariance and transferability across datasets. This ability ultimately translates into improved generalization when compared to the regularization of multiple layers.

\subsection{Qualitative Analysis}
We provide qualitative examples comparing DS-MSHViT and baseline MSHViT predictions using ResNet-101 backbones. The DS-MSHViT matches almost all classes correctly in cases where MSHViT misclassifies some or all defects.
As shown in Fig.~\ref{fig:qual-example} (top left), MSHViT incorrectly predicts no deficiencies, while DS-MSHViT correctly identifies surface damage (OB), displaced joint (FS), branch pipe (GR) and construction changes (OK), only missing the collapse (RB) class. We see similar patterns in the other examples.
In a per-class analysis on the SewerML validation set, DS-MSHViT with ResNet-101 significantly improves performance on RB (cracks, breaks, and collapses) by 3.896\%, DE (deformation) by 3.271\%, IS (infiltration) by 3.295\%, RO (roots) by 3.986\%, BE (attached deposits) by 5.458\%, and FO (obstacles) by 3.547\%. The only class where DS-MSHViT underperforms is chiseled connection (PB) by 3.534\%, most likely due to visual similarity with the chiseled connection (PH) class.
Furthermore, DS-MSHViT overcomes MSHViT's limitations in missing small fine roots in joints, likely because it focuses on the more prevalent displaced joint (FS) and surface damage (OB). MSHViT also missed the RB class which visually resembles the displaced joint (FS) deeper in the pipe.
Overall, the qualitative results demonstrate the advantages of our proposed DS-MSHViT model.

\section{Conclusion}\label{Section:Conclusion}
In this work, we proposed the Dual-Stream Multi-Scale Hybrid Vision Transformer (DS-MSHViT) for automated sewer defect classification.
DS-MSHViT processes RGB and motion modalities through parallel MSHViT streams with an attention consistency loss. By aligning RGB and motion attention maps, DS-MSHViT achieves more accurate and consistent attention for improved classification. The RGB network branch is used to predict the test and validation datasets during inference. Thus, the overall runtime overhead remains identical to the baseline MSHViT, but the RGB branch achieves far superior performance due to the joint multi-modal training.
Experiments show our models outperform MSHViT baselines with different backbones across metrics like F1-Normal, F2-CIW, mAP, OP, and OR. Cross-dataset evaluation on a novel dataset further demonstrates the superior generalization of DS-MSHViT. Qualitative and quantitative analysis highlight the accuracy advantages of DS-MSHViT, which can efficiently exploit visual and motion cues for defect classification. 
Future work involves extending DS-MSHViT capabilities to detection and tracking tasks.
\section{Acknowledgement}
This research was partly supported by the U.S. Department of the Army \textendash U.S. Army Corps of Engineers (USACE) under contract W912HZ-23-2-0004. The views expressed in this paper are solely those of the authors and do not necessarily reflect the views of USACE. 
Authors greatly appreciate Murtada Moussa for providing the Carencro city dataset and Johny Lopez for his work in annotating the Carencro city dataset.
This work has been submitted to the IEEE for possible publication. Copyright may be transferred without notice, after which this version may no longer be accessible.
\bibliographystyle{IEEEtran}  
\bibliography{ref, sewer_ml, vit}
\end{document}